\documentclass[10pt,twocolumn,letterpaper]{article}

\usepackage{3dv}
\usepackage{times}
\usepackage{epsfig}
\usepackage{graphicx}
\usepackage{amsmath}
\usepackage{amssymb}

\usepackage{booktabs}
\usepackage{multirow}
\usepackage{multicol}
\usepackage{times}
\usepackage{epsfig}
\usepackage{graphicx}
\usepackage{amsmath}
\usepackage{amssymb}
\usepackage[inline]{enumitem}
\usepackage{bm}
\usepackage{rotating}
\usepackage{gensymb}
\usepackage{nicefrac}
\usepackage[dvipsnames]{xcolor}
\usepackage{calc}
\usepackage{microtype}  
\usepackage{cuted}  
\usepackage{cite}
\usepackage[bottom]{footmisc}
\usepackage{xr}  
\usepackage{hyperref} 

\usepackage{bbm}
\usepackage{algorithm}
\usepackage{algpseudocode}

\threedvfinalcopy 

\makeatletter
\newcommand{\printfnsymbol}[1]{%
	\textsuperscript{\@fnsymbol{#1}}%
}
\makeatother

\def\threedvPaperID{5} 

\ifthreedvfinal\fi
\begin{document}

\title{Visual Localization via Few-Shot Scene Region Classification}

\author{
Siyan Dong$^{1,3}$\thanks{Joint first authors} \quad Shuzhe Wang$^{2,3}$\printfnsymbol{1} \quad Yixin Zhuang$^{4}$ 
\\
Juho Kannala$^{2}$ \quad Marc Pollefeys$^{3,5}$ \quad Baoquan Chen$^{6}$ 
\\
$^1${Shandong University} \quad $^2${Aalto University} \quad $^3${ETH Zurich} 
\\
$^4${Fuzhou University} \quad $^5${Microsoft} \quad $^6${Peking University}
}

\maketitle
\thispagestyle{empty}

\begin{abstract}

Visual (re)localization addresses the problem of estimating the 6-DoF (Degree of Freedom) camera pose of a query image captured in a known scene, which is a key building block of many computer vision and robotics applications. Recent advances in structure-based localization solve this problem by memorizing the mapping from image pixels to scene coordinates with neural networks to build 2D-3D correspondences for camera pose optimization. However, such memorization requires training by amounts of posed images in each scene, which is heavy and inefficient. On the contrary, few-shot images are usually sufficient to cover the main regions of a scene for a human operator to perform visual localization. In this paper, we propose a scene region classification approach to achieve fast and effective scene memorization with few-shot images. Our insight is leveraging a) pre-learned feature extractor, b) scene region classifier, and c) meta-learning strategy to accelerate training while mitigating overfitting. We evaluate our method on both indoor and outdoor benchmarks. The experiments validate the effectiveness of our method in the few-shot setting, and the training time is significantly reduced to only a few minutes.\footnote{Code available at: \url{https://github.com/siyandong/SRC}}
   
\end{abstract}


\section{Introduction}

Visual (re)localization is a key component of many computer vision and robotics applications such as Augmented Reality (AR) and navigation. It addresses the problem of estimating the 6-DoF (Degree of Freedom) camera pose of a query image captured in a known scene. There are mainly two types of approaches: direct pose estimation by image retrieval \cite{sattler2012image, arandjelovic2013all, torii201524, arandjelovic2016netvlad} or pose regression \cite{kendall2015posenet, Kendall_2017_CVPR, walch2017image, sattler2019understanding, wang2020atloc}, and two-step pose estimation (also known as structure-based localization) \cite{sattler2016efficient, dusmanu2019d2, sarlin2019coarse, sarlin2020superglue, sarlin2021back, li2018full, li2020hierarchical, wang2021continual, brachmann2017dsac, brachmann2018lessmore, brachmann2021dsacstar, huang2021vs, yang2019sanet, tang2021learning} that first infers correspondences between image pixels and scene points and then solves camera pose by geometric optimization. During the last decade, scene coordinate regression based two-step approaches \cite{li2020hierarchical, brachmann2021dsacstar, huang2021vs, tang2021learning} achieve state-of-the-art localization accuracy on public benchmarks~\cite{shotton2013scene,valentin2016learning,kendall2015posenet}, which is the main focus of this paper.

\begin{figure}[t]
\centering
	\includegraphics[width=0.9\linewidth]{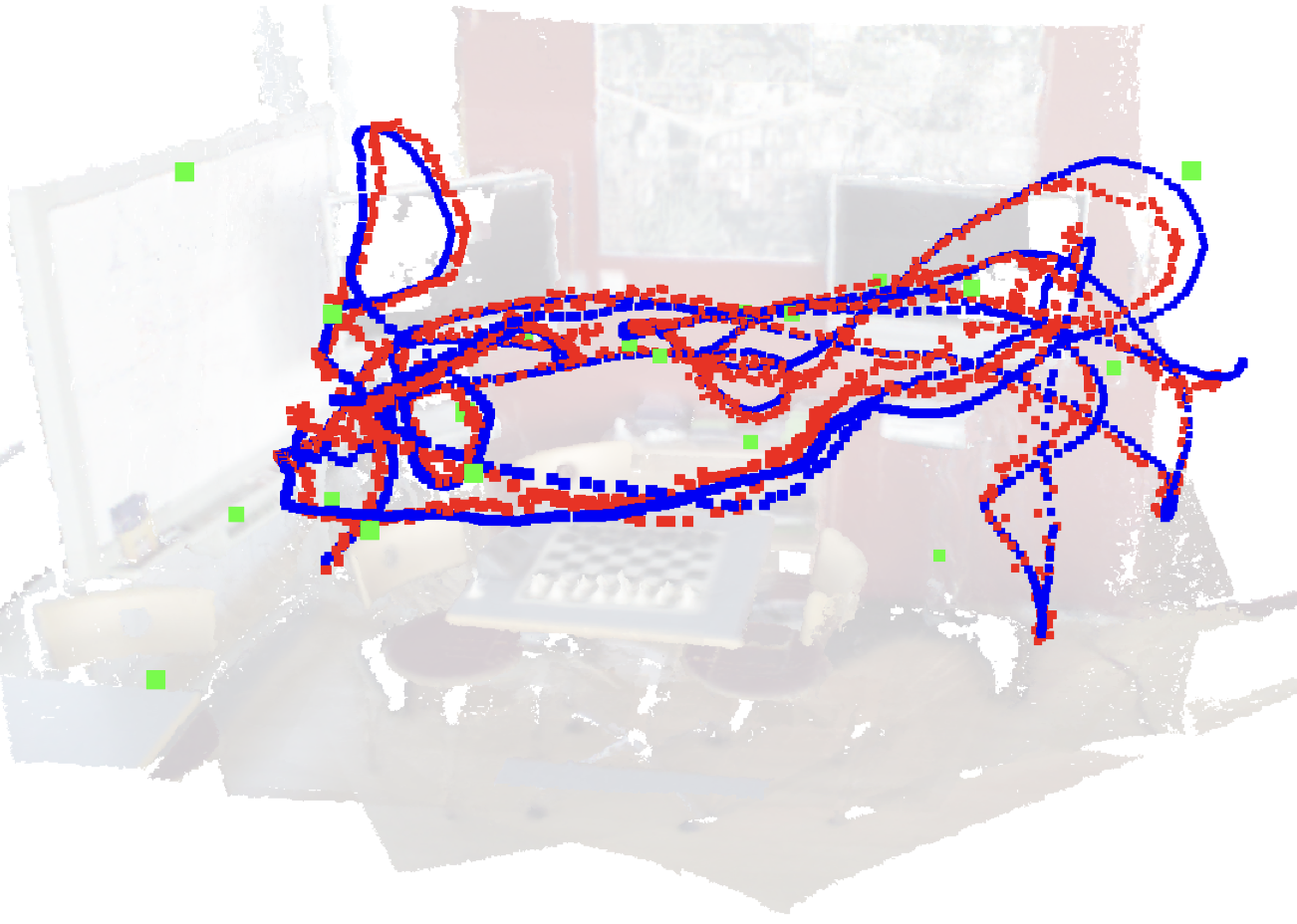}
	\caption{Visualization of our estimated (red dots) and ground truth (blue dots) camera poses in the scene \textsc{Chess}~\cite{shotton2013scene}. Our method is trained with 20 images (green dots) uniformly sampled from the original 4000 training images. We observe that the 20 images are enough to cover the main regions of the scene and construct a coarse 3D model (shown in the background) to support our method.}
	\label{figure:teaser}
\end{figure}

The essential concept of scene coordinate regression is to memorize the mapping from image pixels to scene coordinates under a variety of viewpoints. A common way is to leverage convolutional neural networks (CNNs) and encode the aforementioned mapping as the network parameters. As a result, most of them require amounts of posed images to train their models in each specific scene, 
therefore can hardly be deployed in practice.
On the contrary, few-shot images are usually sufficient to cover the main regions of a scene, as shown in Figure \ref{figure:teaser}, and sufficient for a human operator to memorize a scene for future localization purposes. One hypothesis is that we humans have learned during daily life the prior knowledge to extract robust visual features, and only have to memorize the locations of different features in a novel scene. 
Incorporating the aforementioned insights into the scene coordinate regression based method, in this paper, we study the fast memorization of novel scenes with only few-shot posed images as training data. 

Introducing pre-learned visual feature extractors \cite{detone2018superpoint, dusmanu2019d2, revaud2019r2d2, sun2021loftr} is not new in the visual localization community. Especially, as another popular series of the two-step pose estimation, the feature matching based methods make use of off-the-shelf feature detectors and descriptors to build scene models from posed images, to match query image features and scene coordinates. They are in nature generalized to different scenes as the visual features are scene-agnostic. 
However, they usually cost large memory and are not accurate as the scene-specific coordinate regression based methods.
Therefore, we propose a new paradigm that combines scene-agnostic features and scene-specific coordinate estimation. 

Our key idea is that the scene-agnostic feature extractor is helpful for fast scene memorization, while the scene-specific training also helps with accuracy and efficiency. Such insight motivates us to decouple the popular end-to-end scene coordinate regression pipeline into a scene-agnostic feature extractor and a scene-specific coordinate estimator. Further, we propose a scene region classification paradigm instead of direct coordinate regression for coordinate inference, so that a meta-learning strategy is applied to accelerate training.

We summarize our contributions as follows:
\begin{itemize}

\item We propose a novel problem setting, i.e. visual localization with only few-shot posed images as the database, along with a simple and effective method designed for the proposed few-shot setting. 

\item Leveraging both scene-agnostic and scene-specific information, we introduce scene region classification and meta-learning strategy for fast scene memorization. 

\item Experiments validate the effectiveness of our method. In the few-shot setting, we outperform the state-of-the-art scene coordinate regression based methods, with only a few minutes of training time. 

\end{itemize}


\section{Related Work}

\textbf{Direct pose estimation.} This type of approach directly estimates the camera pose based on the query image. An effective direction is based on image retrieval methods \cite{sattler2012image, arandjelovic2013all, torii201524, arandjelovic2016netvlad}. They aggregate either hand-crafted or learned local features into whole-image global descriptors to match the query image against the database. PoseNet \cite{kendall2015posenet} and its variants \cite{Kendall_2017_CVPR, walch2017image, sattler2019understanding, wang2020atloc} make use of neural networks to directly regress camera pose. 
There are also RGB-D based methods \cite{glocker2014real, dai2017bundlefusion, uy2018pointnetvlad} that encode both color and depth information into global features. Although they are efficient, the localization results are not as accurate as the two-step pose estimation methods. Therefore, the approaches of direct pose estimation are not compared in this paper.

\textbf{Two-step pose estimation.} This type of approach first infers correspondences between image pixels and scene points and then solves camera pose by optimization. To obtain the aforementioned 2D-3D correspondences, conventional feature matching based methods \cite{sattler2016efficient, dusmanu2019d2, sarlin2019coarse, sarlin2020superglue, sarlin2021back} leverage either hand-crafted or learned keypoint features to explicitly construct scene maps to match the query image's keypoints. Another popular direction is scene coordinate regression based methods \cite{li2018full, li2020hierarchical, wang2021continual, brachmann2017dsac, brachmann2018lessmore, brachmann2021dsacstar, huang2021vs} that estimate the 2D-3D correspondences implicitly by memorizing the mapping from image pixels to scene coordinates. 
Such scene-specific memorization designs usually achieve more accurate results than scene-agnostic feature matching, while they need to be trained for each novel scene.
Recent works \cite{yang2019sanet, tang2021learning} propose to regress scene coordinate based on retrieved training image coordinate.
There are also RGB-D based visual localization approaches  \cite{shotton2013scene, valentin2015exploiting, cavallari2017fly, cavallari2019real, dong2021robust} leveraging random forests or point cloud based backbones, with 3D-3D correspondence optimization algorithms, which is beyond our scope.

\textbf{Few-shot learning.} Few-shot learning, also known as low-shot learning, is the learning paradigm with only a few training data. There are roughly three types of approaches: data hallucination based methods \cite{hariharan2017low, antoniou2017data, wang2018low} that leverage generators to augment training data, metric learning based methods \cite{koch2015siamese, vinyals2016matching, snell2017prototypical} that learn feature representations for similarity comparisons, and meta-learning based methods \cite{andrychowicz2016learning, finn2017model, nichol2018reptile} that aim to find a proper initialization for fast training on a new task. 
In this paper, we adopt the meta-learning strategy and obtain the initial network parameters to perform fast memorization of novel scenes.

\section{Method}

\begin{figure*}[t]
\centering
	\includegraphics[width=0.95\linewidth]{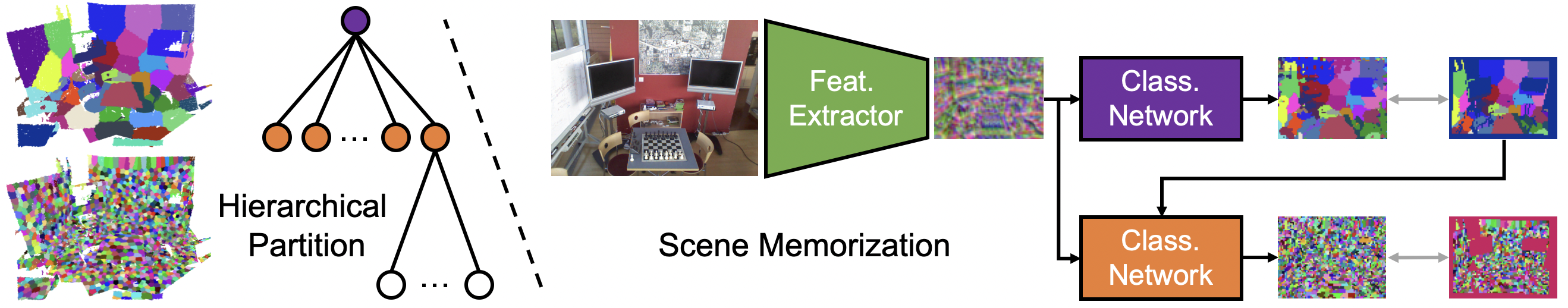}
	\caption{The training pipeline of our method. A hierarchical partition tree is built to divide the scene to regions, and then a neural network is trained to map input image pixels to region labels. The network is designed to leverage both scene-agnostic priors (i.e. the feature extractor SuperPoint~\cite{detone2018superpoint}) and scene-specific memorization (i.e. the classifier that consists of hierarchical classification networks).}
	\label{figure:train}
\end{figure*}

\textbf{Problem statement.} Visual localization is a scene-specific problem, with the requirement of a scene-specific database as the reference for camera pose estimation. Specifically, in each scene, a set of posed RGB-D images are given as the database to estimate the 6 DoF camera pose of single query RGB images. Different from the previous settings \cite{li2020hierarchical, brachmann2021dsacstar}, there are only a few rather than thousands of training images given in our problem. The assumption is that tens of images are sufficient to cover the major regions of a scene to support visual localization.

\begin{figure*}[t]
\centering
	\includegraphics[width=0.95\linewidth]{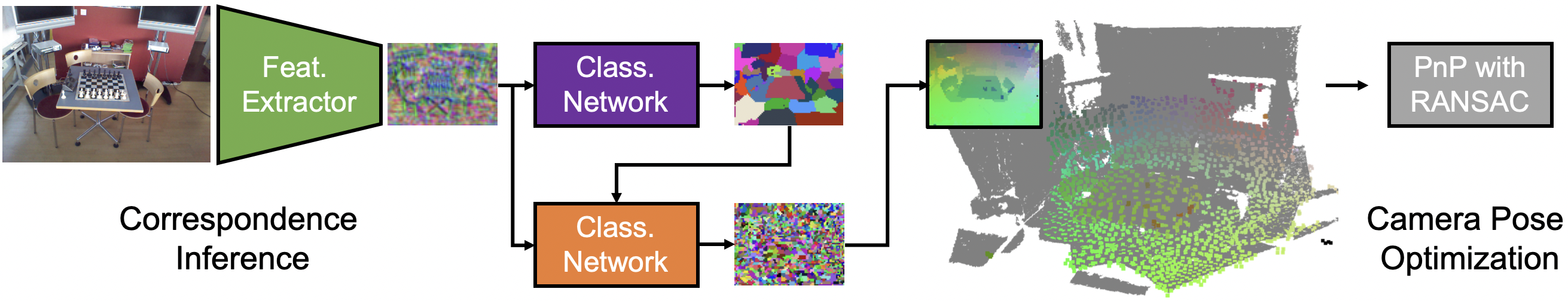}
	\caption{The camera pose estimation pipeline of our method. Given a query image, the trained network infers correspondences between image pixels and scene regions. Since each scene region corresponds to a set of scene coordinates, 2D-3D correspondences are built between image pixels and scene coordinates. Followed by a PnP algorithm with RANSAC, the camera pose is solved by optimization.}
	\label{figure:infer}
\end{figure*}

\textbf{Method overview.} Our method follows the two-step pose estimation framework and is a variant of scene coordinate regression. 
The framework first builds the 2D-3D correspondences between image pixels and scene coordinates and then optimizes the camera poses with a principled geometric algorithm.
Figure \ref{figure:train} illustrates our training pipeline, which adopts a from-agnostic-to-specific structure.
Given the few-shot RGB-D training images, we first build a hierarchical scene partition tree to label each image pixel a set of region IDs. 
Then, instead of directly regressing the scene coordinates, we employ a neural network to map the image pixels to their corresponding region IDs.
Particularly, the network is composed of two components: a scene-agnostic visual feature extractor $f$ (\ref{subsec:extractor}) and a scene-specific region classifier $c$ (\ref{subsec:classifier}).
Figure \ref{figure:infer} illustrates our camera pose estimation pipeline. Through hierarchical classification, an input 2D pixel is mapped to a leaf node, i.e. a region consisting of a compact set of scene coordinates, as its 3D correspondences. Such one-to-many correspondences are fed to a Perspective-n-Point (PnP) \cite{gao2003complete, ke2017efficient} algorithm inside a RANSAC \cite{fischler1981random, sattler2019understanding} loop for camera pose optimization (\ref{subsec:optimization}).

\subsection{Scene-Agnostic Feature Extractor} \label{subsec:extractor}

Recent work \cite{sattler2019understanding} interprets the direct pose regression networks as linear combinations of embedded features. 
In this work, we follow a similar hypothesis to interpret the typical scene coordinate regression network SCRNet~\cite{li2020hierarchical}: 
the first-to-middle layers serve as a descriptor extractor and the rest layers serve as a coordinate regressor. 
The reason for requiring amounts of training data is that the visual descriptors should be learned to be robust under different viewpoints
(please refer to \ref{subsec:analysis} scene coordinate regression). 
Since recently learned descriptors \cite{detone2018superpoint, dusmanu2019d2, revaud2019r2d2, sun2021loftr} achieve promising visual feature embedding ability under viewpoint changes and cross multiple scenes, we make use of the off-the-shelf SuperPoint~\cite{detone2018superpoint} (without detector) as our feature extractor, so that we can obtain robust semi-dense feature maps from an input image. Such a feature extractor $f$ is scene-agnostic and can be directly deployed in novel scenes without any training. 
Specifically, the feature extractor $f$ is a VGG-Style \cite{simonyan2014very} network that inputs an image $I_{H \times W}$ and outputs its feature map $F_{H_o \times W_o} = f(I_{H \times W})$, where $H_o = H/8$ and $W_o = W/8$. The output features are regarded as robust descriptors for $8 \times 8$ image patches and serve as the input for our scene-specific region classifier.

\subsection{Scene-Specific Region Classifier} \label{subsec:classifier}
What makes visual localization a scene-specific problem is memorizing the scene-specific coordinate system.
Direct coordinate regression takes time to converge even with only few-shot training images (please refer to \ref{subsec:analysis} training times).
Instead, in this subsection, we introduce a hierarchical scene region classification approach for fast memorization of the mapping from image pixels to scene regions. 

\textbf{Scene partition tree.} The goal of the scene partition is to divide scene coordinates into clusters in order to convert the task of coordinate regression to region classification. Given a set of RGB-D training images, we first fuse a scene point cloud according to their depth images and camera poses. On top of the scene point cloud, we build a hierarchical partition tree. Specifically, we run the classic K-Means algorithm to obtain $m$ clusters for the first level of region partition. By iteratively executing region partition for each cluster, we eventually obtain a $n$-level $m$-way tree. Each node in the tree corresponds to a specific scene region, and each image pixel is automatically labeled by hierarchical node IDs. 

\textbf{Hierarchical classification networks.} The classifier $c$ aims to map a pixel to a leaf node. It consists of hierarchical classification networks, each of which performs a $m$-class classification. The whole process is denoted by $P = c(f(I))$, where $P \in \{1,2,...,m^n\}_{H_o \times W_o}$ denotes the final classification map for the input feature map. Specifically, each level $l$ in the tree corresponds to a classification network $c_l$. For the first level (the root node), the network $c_1$ takes the image feature map as input and outputs $m$-class probabilities: 
\begin{equation}
P_1 = c_1(F), 
\end{equation}
where $P_1 \in (0,1)_{m \times H_o \times W_o}$ denotes the classification probability map.
For the rest levels, each network $c_l$ inputs both image feature map and predicted classification probability from previous levels:
\begin{equation}
P_l = c_{l}(F, P_{1,...,l-1}),
\end{equation}
where $P_l \in (0,1)_{m \times H_o \times W_o}$. The final classification for the leaf nodes is computed by 
\begin{equation}
P = \sum_{i=1}^n a(P_i)*m^{n-i},
\end{equation}
where $a(\cdot)$ denotes the argmax operator at the first channel to convert the one-hot probability to a class label.

\begin{figure}[tb]
\centering
	\includegraphics[width=0.95\linewidth]{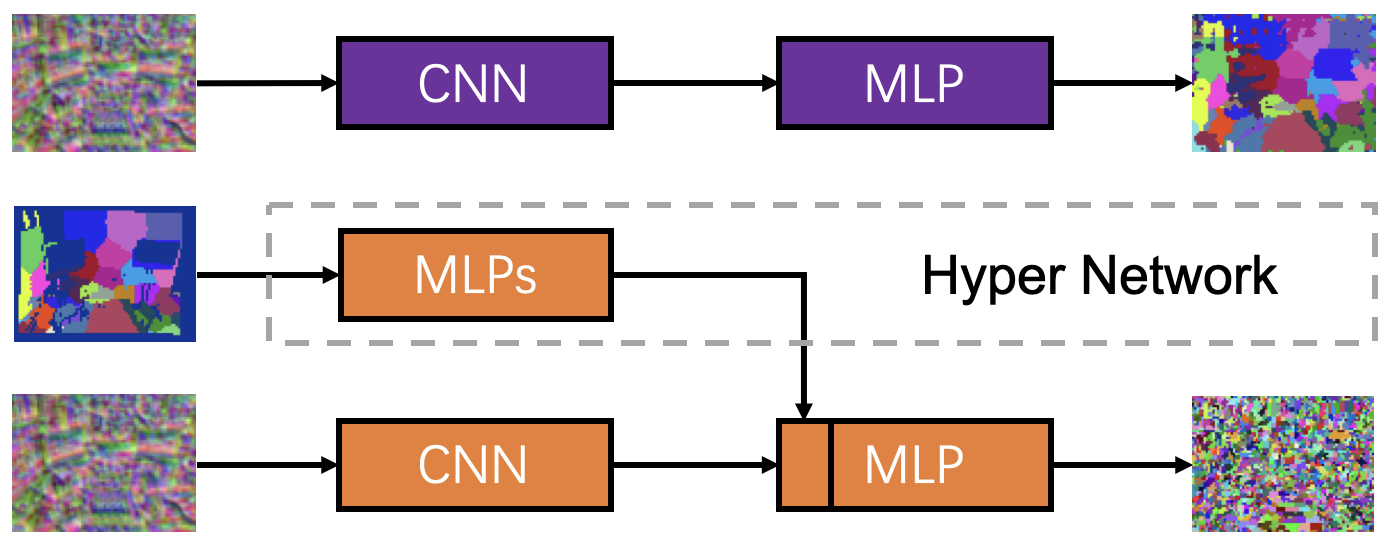}
	\caption{Illustration of the scene-specific region classifier. It consists of hierarchical classification networks. For the first level, it is only a base network. For the rest levels, each network contains both a base network and a hyper network. }
	\label{figure:classifier}
\end{figure}

Following recent works \cite{li2020hierarchical, dong2021robust}, the classification networks are implemented as base and hyper networks.
As shown in Figure \ref{figure:classifier}, at the first level, the feature maps from the previous extraction module are fed to a CNN to extract scene-specific feature patterns, which are followed by a pixel-wise MLP to output the classification probability map.  
Starting from the second level, the feature patterns are modulated by a hyper network, according to the classification probability from previous levels.
The intuition of the modulation is that similar feature patterns appearing in different scene regions should be classified under different labels. 
Since different levels handle the feature patterns on different scales, the corresponding CNNs are implemented with different receptive fields. Specifically, we make use of dilated convolutions to control the receptive fields, and the dilation parameters are computed according to the average cluster radius at each level. Please refer to \ref{subsec:details} for implementation details.

\textbf{Training with meta-learning.} 
We supervise the training process at each classification output with a pixel-wise cross-entropy loss. The loss term is formulated as: 

\begin{equation}
\begin{split}
    Loss_{i,j}(P_i^{(j)}, Y_i^{(j)}) = & \\
    - \sum_{d=1}^m \log & \frac {\exp{( P_{i,d}^{(j)} )}} { \sum_{k=1}^m \exp{( P_{i,k}^{(j)} )} } \mathbbm{1} (Y_i^{(j)}=d),
\end{split}
\end{equation}
\vspace{3mm}

where $P_i^{(j)} \in (0,1)_m$ denotes the $m$-class probability prediction of pixel $j$ at level $i$, $Y_i^{(j)} \in \{1,2,..,m\}$ denotes the ground truth label, and $\mathbbm{1}(\cdot)$ denotes a binary function. Note that at the training stage, we use the ground truth label map as the input for each hyper network. Therefore, the multi-level classification network can be trained in parallel at each branch to reduce the training time. 

Inspired by the recent meta-learning idea, we apply Reptile~\cite{nichol2018reptile} strategy to initialize the hierarchy network with a pre-trained model.
The pre-training aims to find proper network parameters so that for a novel scene the network can achieve fast convergence. 
This process makes use of $t$ individual scenes as a set of hierarchical classification tasks $T=\{T_i \vert i=1,2,...,t\}$. The classifier parameters $\phi_{beg}$ are randomly initialized and iteratively updated by the Reptile gradient. For each iteration $i$, a task $T_i$ is randomly sampled from $T$ to perform $k$ steps of stochastic gradient descent (SGD) operation, starting with $\phi_{i-1}$, and resulting in $\tilde{\phi}_{i-1}$. The Reptile gradient $G_i$ is computed as 
\begin{equation}
    G_i = \tilde{\phi}_{i-1} - \phi_{i-1}.
\end{equation}
Then, the network parameters are actually updated as 
\begin{equation}
    \phi_{i} = \phi_{i-1} + \epsilon G_i.
\end{equation}
The iteration step is repeated until convergence. Although there may be conflict among different tasks, Reptile gradients make the training process converge and we obtain $\phi_{end}$. Then in each few-shot scene, we apply $\phi_{end}$ as our initial network parameters to perform fast memorization. 



\subsection{Camera Pose Optimization} \label{subsec:optimization}

Our localization pipeline leverages the trained network to build 2D-3D correspondences between image pixels and scene regions to perform the camera pose optimization. Each pixel is mapped to a leaf node in the scene partition tree, which corresponds to a specific scene region with a set of 3D coordinates. To limit the number of coordinates, we further partition each leaf node to $q$ clusters by the K-Means algorithm and utilize the cluster centers for final pose estimation. Therefore, a 2D pixel $x_i \in \mathbb{N}^2$ is corresponded to a set of $q$ 3D coordinates $X_i = \{X_{i,k} \in \mathbb{R}^3 \vert k=1,2,...,q\}$.

Let $M=\{(x_i, X_i) \vert i=1,2,...,N \}$ denote the set of 2D-3D correspondences.
In contrast to the recent works~\cite{brachmann2018lessmore, li2020hierarchical}
for applying the PnP solver, our camera pose optimization is a variant of PnP-RANSAC algorithm to handle one-to-many correspondences. 
Our algorithm consists of three stages, which are introduced in detail below.


\textbf{Hypotheses Generation.} The goal is to generate a set of camera pose hypothesis $H=\{H_i \in \mathbf{SE}(3) \vert i=1,2,...,h\}$. 
We first randomly sample 4 correspondences from $M$. For each sampled pixel $x_i$, we randomly select a 3D coordinate $X_{i,k}$ from $X_i$ to form a one-to-one match. The 4 one-to-one matches formulate a minimal set for a PnP algorithm \cite{gao2003complete, ke2017efficient} to solve a unique camera pose. This process is executed $h$ times to obtain the hypotheses set $H$.

\textbf{Ranking.} We then select the hypothesis that reaches the most consensus with the input correspondences $M$. For each generated hypothesis $H_i$, we first compute the reprojection error for each correspondence indexed by $j$ as 
\begin{equation} \label{eq:reproj}
    e_j(H_i, M_j) = \min\nolimits_{k=1}^q \lVert x_j - Proj(H_i X_{j,k}) \rVert_2.
\end{equation}

Then we apply a kernel function to the pixel-wise reprojection errors and  sum them to obtain the consensus score:

\begin{equation}
\begin{split}
    S_i(H_i, M) = & \\
    \sum_{j=1}^N & \frac{1}{ 1+ \exp (-0.5*( e_j(H_i, M_j) - \tau )) } , 
\end{split}
\end{equation}

where $\tau=10$ is a threshold in the pixel coordinate system. We sort all the hypotheses by their scores and select the one with the lowest score as the output camera pose.

\textbf{Refinement.} This is the post-processing of the selected hypothesis to obtain a fine-grained camera pose as the final output. The refinement is executed iteratively. In each iteration, we recompute the reprojection error and select inlier correspondences whose error is less than the threshold $\tau$. Then these inliers are fed into another PnP solver with Levenberg-Marquardt optimization \cite{eade2013gauss} to refine the camera pose. The iteration terminates when the camera pose converges, or until the iterations reach a max limit of $20$ steps.

\section{Experiments}
In this section, we validate the effectiveness of our method. We first introduce the few-shot version of datasets (\ref{subsec:dataset}), and describe implementation details (\ref{subsec:details}). Then we perform comparisons with state-of-the-art methods (\ref{subsec:comp}). Last but not the least, we analyze our method against our motivation and conduct several ablation studies (\ref{subsec:analysis}).

\subsection{Datasets}\label{subsec:dataset}
We evaluate our method on two standard benchmarks, 7-Scenes dataset~\cite{shotton2013scene} and Cambridge landmarks~\cite{kendall2015posenet}. 
The 7-Scenes dataset consists of 7 indoor scenes with RGB-D images while the Cambridge landmarks contain 6 outdoor scenes. 
As the scene \textsc{Street} in Cambridge failed to provide accurate 3D reconstruction, following the previous work~\cite{li2020hierarchical, brachmann2021dsacstar}, we conduct experiments only on the rest 5 scenes. 
Besides the two datasets, we leverage the 12-Scenes dataset~\cite{valentin2016learning} to pre-train our classification network with Reptile~\cite{nichol2018reptile} for model initialization.

In our few-shot setting, we uniformly sample $0.5\sim1.0\%$ images from the original training set. The selected training numbers are presented in Table~\ref{tab:main1} and Table~\ref{tab:main2}. 
\begin{table*}[th]
\centering

\renewcommand{\b}[1]{\textbf{#1}}
\Large
\renewcommand\arraystretch{1.5}
\resizebox{0.98\textwidth}{!}{
\begin{tabular}{lccccccccccccc}
\hline
{\multirow{2}{*}{Methods}} & \multicolumn{6}{c}{Original training (median pose error in cm/$^\circ$) }             &            & \multicolumn{6}{c}{Few-shot training (median pose error in cm/$^\circ$) }                               \\ \cline{2-7} \cline{9-14}
& \# Images & \textcolor{orange}{AS}~\cite{sattler2016efficient}$^\dag$    & \textcolor{orange}{HLoc}~\cite{sarlin2019coarse,sarlin2020superglue}   & SCRNet~\cite{li2020hierarchical} & HSCNet~\cite{li2020hierarchical} & DSAC*~\cite{brachmann2021dsacstar}  && \# Images & \textcolor{orange}{HLoc}~\cite{sarlin2019coarse,sarlin2020superglue}   & DSAC*~\cite{brachmann2021dsacstar}  & HSCNet~\cite{li2020hierarchical}  & \textbf{SP+Reg}    & \textbf{Ours}   \\ \cline{1-7} \cline{9-14}
     \textsc{Chess} & 4000 & 
     3/0.87 & 2/0.85 & 2/0.7 & 2/0.7 & 2/1.10 &
     & 20 & 
     \textcolor{blue}{4}/1.42 & \textcolor{red}{3/1.16} & \textcolor{blue}{4}/1.42 & \textcolor{blue}{4}/1.28 & \textcolor{blue}{4/1.23} \\
     
     \textsc{Fire} & 2000 & 
     2/1.01 & 2/0.94 & 2/0.9 & 2/0.9 & 2/1.24 & 
     & 10 & 
     \textcolor{red}{4}/1.72 & \textcolor{blue}{5}/1.86 & \textcolor{blue}{5/1.67} & \textcolor{blue}{5}/1.95 & \textcolor{red}{4/1.53} \\
     
     \textsc{Heads} & 1000 & 
     1/0.82 & 1/0.75 & 1/0.8 & 1/0.9 & 1/1.82 & 
     & 10 & 
     4/\textcolor{blue}{1.59} &
     4/2.71 & 
     \textcolor{blue}{3}/1.76 & \textcolor{blue}{3}/2.05 & \textcolor{red}{2/1.56} \\
     
     \textsc{Office} & 6000 & 
     4/1.15 & 3/0.92 & 3/0.9 & 3/0.8 & 3/1.15 & 
     & 30 & 
     \textcolor{red}{5/1.47} & 
     9/2.21 & 
     9/2.29 & 
     \textcolor{blue}{7/1.96} & \textcolor{red}{5/1.47} \\
     
     \textsc{Pumpkin} & 4000 & 
     7/1.69 & 5/1.30 & 4/1.1 & 4/1.0 & 4/1.34 & 
     & 20 & 
     \textcolor{blue}{8}/\textcolor{blue}{1.70} & 
     \textcolor{red}{7/1.68} & 
     \textcolor{blue}{8}/1.96 & 
     \textcolor{red}{7}/1.77 & \textcolor{red}{7}/1.75 \\
     
     \textsc{RedKitchen} & 7000 & 
     5/1.72 & 4/1.40 & 5/1.4 & 4/1.2 & 4/1.68 & 
     & 35 & 
     \textcolor{blue}{7}/\textcolor{red}{1.89} &
     \textcolor{blue}{7}/2.02 &
     10/2.63 & 
     8/2.19 & \textcolor{red}{6}/\textcolor{blue}{1.93} \\
     
     \textsc{Stairs} & 2000 & 
     4/1.01 & 5/1.47 & 4/1.0 & 3/0.8 & 3/1.16 & 
     & 20 & 
     \textcolor{blue}{10/2.21} & 
     18/4.8 & 
     13/4.24 & 
     120/27.37 & 
     \textcolor{red}{5/1.47} \\ \hline
\end{tabular}}
\vspace{6pt}
\caption{Visual localization results on the 7-Scenes dataset. Left: original training images as database. Right: the few-shot training images as database. For each part of the results, we first list the nunmbers of training images in each scene. The feature matching based methods are labeled to orange. The best and the second best results among the few-shot setting are labeled to red and blue, respectively.}
\vspace{10pt}

\renewcommand{\b}[1]{\textbf{#1}}
\renewcommand\arraystretch{1.5}
\Large
\resizebox{0.98\textwidth}{!}{
\begin{tabular}{lccccccccccccc}
\hline
{\multirow{2}{*}{Methods}} & & \multicolumn{6}{c}{Original training (median pose error in cm/$^\circ$) }                            & \multicolumn{6}{c}{Few-shot training (median pose error in cm/$^\circ$) }                                   \\ \cline{2-7} \cline{9-14}
                        & \# Images & \textcolor{orange}{AS}~\cite{sattler2016efficient}$^\dag$    & \textcolor{orange}{HLoc}~\cite{sarlin2019coarse,sarlin2020superglue}   & SCRNet~\cite{li2020hierarchical} & HSCNet~\cite{li2020hierarchical} & DSAC*~\cite{brachmann2021dsacstar}  && \# Images & \textcolor{orange}{HLoc}~\cite{sarlin2019coarse,sarlin2020superglue}   & DSAC*~\cite{brachmann2021dsacstar}  & HSCNet~\cite{li2020hierarchical}  & \textbf{SP+Reg}    & \textbf{Ours}   \\ \cline{1-7} \cline{9-14}
                            \textsc{GreatCourt}     & 1531      & 24/0.13 & 16/0.11 & 125/0.6 & 28/0.2 & 49/0.3 & &16        & \textcolor{red}{72/0.27} & NA       & NA       & NA       & \textcolor{blue}{81/0.47} \\
                            \textsc{KingsCollege}   & 1220      & 13/0.22 & 12/0.20 & 21/0.3  & 18/0.3 & 15/0.3 && 13        & \textcolor{red}{30/0.38} & 156/2.09 & 47/0.74  & 111/1.77 & \textcolor{blue}{39/0.69} \\
                            \textsc{OldHospital}    & 895       & 20/0.36 & 15/0.30 & 21/0.3  & 19/0.3 & 21/0.4 & &9         & \textcolor{red}{28}/\textcolor{blue}{0.42} & 135/2.21 & \textcolor{blue}{34}/\textcolor{red}{0.41}  & 116/2.55 & 38/0.54 \\
                            \textsc{ShopFacade}     & 229       & 4/0.21  & 4/0.20  & 6/0.3   & 6/0.3  & 5/0.3  && 3         & 27/1.75  & NA       & \textcolor{blue}{22/1.27}  & NA       & \textcolor{red}{19/0.99} \\
                            \textsc{StMarysChurch}  & 1487      & 8/0.25  & 7/0.21  & 16/0.5  & 9/0.3  & 13/0.4 & &15        & \textcolor{red}{25/0.76} & NA       & 292/8.89 & NA       & \textcolor{blue}{31/1.03} \\ \hline
\end{tabular}} \label{tab:main1}
\vspace{6pt}
\caption{Visual localization results on the Cambridge landmarks. NA indicates that the method fails (median translation error larger than 500cm) in the scene. The results of AS$^\dag$ come from the paper PixLoc~\cite{sarlin2021back}. DSAC* is trained with 3D models.
HLoc uses SuperPoint for keypoint detection \& description, and SuperGlue~\cite{sarlin2020superglue} for feature matching.
}
\label{tab:main2}
\end{table*}

\subsection{Implementation Details}\label{subsec:details}
For all the scenes, we apply the hierarchical K-Means algorithm \cite{scikit-learn} and implement our scene partition as $2$-level trees. Therefore, each scene-specific region classifier consists of two levels of classification networks. By default, we set $m=64$ for 7-Scenes dataset, and $m=100$ for Cambridge landmarks. Therefore, each indoor scene is divided into $4096$ regions, and each outdoor scene is divided into $10000$ regions.
To construct the partition tree, we make use of the few-shot depth images and their camera poses to fuse the scene point cloud. 
Specifically, we use original depth images for 12-Scenes, calibrated depth images for 7-Scenes, and rendered depth images for Cambridge. The calibrated and rendered depth images are provided by DSAC++ \cite{brachmann2018lessmore}.

The input image resolution to our feature extractor $f$ is fixed to $480 \times 640$ (default resolution of 12-Scenes and 7-Scenes datasets). The images from Cambridge landmarks are resized to $ 480 \times 852 $ and randomly cropped to $480 \times 640$ for training, and centered cropped for inference. The output feature map from extractor $f$ is of shape $256 \times 60 \times 80$, where $256$ is the dimension of the features. Then, the extracted map is fed to the scene-specific region classifier $c$ and outputs the region predictions with the shape of $m \times 60 \times 80$. 
Each classification module consists of a $2$-layer CNN and a $2$-layer MLP. The dilation of each Convolution layer is set to $5$ for the first level, $3$ for the second level. The hyper network consists of two $2$-layer MLPs to generate feature normalization parameters $\gamma$ and $\beta$ ($\gamma, \beta \in \mathbb{R}^{256 \times 60 \times 80}$).
The feature modulation is implemented as 
\begin{equation}
    F_{out} = \gamma * F_{in} + \beta.
\end{equation}
Except for the final output layer, each layer is followed by a layer normalization \cite{ba2016layer, wu2018group} and a ReLU activation \cite{agarap2018deep}. 

We set the batch size to $1$ and the learning rate to $5e-4$ in all of the training procedures. For Reptile pre-training on the 12-Scenes dataset, the number of tasks $t$ is set to $12$. We use $k=2$ SGD steps, and $\epsilon=5e-4$. For the fast memorization on 7-Scenes and Cambridge, we apply the Adam optimization algorithm \cite{kingma2014adam}. As for inference, the number of leaf coordinates is limited to $q=10$ for both 7-Scenes and Cambridge. The number of hypotheses is set to $h=256$ for 7-Scenes, and $h=512$ for Cambridge.
All the experiments are run on NVIDIA GeForce RTX 2080 Ti GPU and AMD Ryzen Threadripper 2950x CPU.

\subsection{Comparison} \label{subsec:comp}
We compare our method with state-of-the-art two-step pose estimation methods. We consider both feature matching based methods (AS~\cite{sattler2016efficient} and HLoc~\cite{sarlin2019coarse,sarlin2020superglue}) and scene coordinate regression based methods (SCRNet~\cite{li2020hierarchical}, HSCNet~\cite{li2020hierarchical}, and DSAC*~\cite{brachmann2021dsacstar}) as our competitors. We also set up a baseline, named SP+Reg, that equips SuperPoint~\cite{detone2018superpoint} as a feature extractor while employing the regression network instead of classification for the coordinate estimation. 

\textbf{Quantitative results.} The quantitative results are shown in Table~\ref{tab:main1} and Table~\ref{tab:main2}. 
For each dataset, we list the localization results trained by the original training set in the left part, as the reference to compare with results in the few-shot setting. We first compare the results between the original and the few-shot training. All of the methods suffer an obvious performance drop as the few-shot setting is challenging. The results of scene coordinate regression based methods on Cambridge landmarks are overall worse than the feature matching based ones, as the ground truth depth images used for training are rendered from 3D reconstruction and not as accurate as 7-Scenes measured depth images.

When compared with the scene coordinate regression based methods in the few-shot setting, our method achieves the best performance on both the 7-Scenes dataset and Cambridge landmarks. 
Note that for fair comparisons, all SCRNet, HSCNet, SP+Reg, and our method are trained with 9K iterations for 7-Scenes and 30K for Cambridge to ensure the aforementioned methods converge. DSAC* is trained with 100K iterations in the initialization step and 20K iterations in the end-to-end training, and the average training time for each scene is around 2 hours. 
Besides, the regression-only methods (DSAC* and SP+Reg) fail in several outdoor scenes, the hybrid method (HSCNet) fails fewer, while the classification-only method (Ours) achieves reasonable results across all the scenes.

We also notice that in the few-shot setting, our method overall falls behind the state-of-the-art method HLoc on the Cambridge landmarks.
It is reasonable since feature matching based methods do not require the rendered depth for supervision. 
Nevertheless, our method consumes low memory storage ($\sim$40 MB) and fast inference time ($\sim$200ms).
In addition, the main purpose of this paper is to analyze the scene coordinate regression architecture and design a simple and effective scene memorization pipeline to increase the generalisability for novel scenes. 
Exploring the method combinations of feature matching and coordinate regression in the few-shot setting is worth more research.

\begin{figure*}[tbh]
\centering
	\includegraphics[width=1\linewidth]{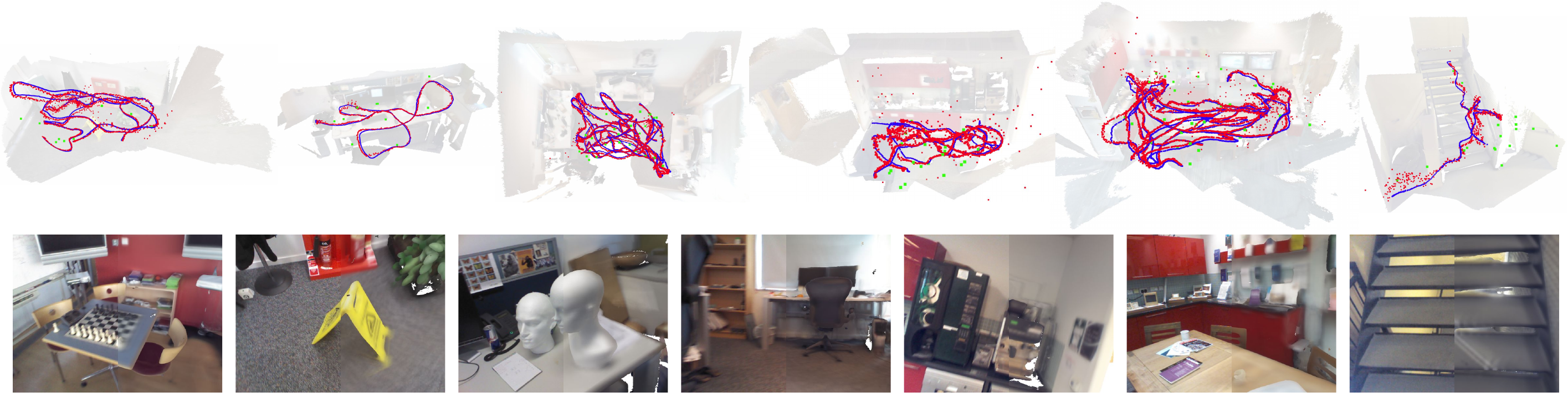}
	\vspace{-4mm}
	\caption{Visual results on the 7-Scenes dataset. Top: we visualize the camera poses of training images (green dots), test images (blue dots) and our estimates (red dots). Bottom:  we select the image with the median pose error in each scene and visualize the accuracy by overlay the query image (left) with a rendered image (right) using the estimated pose and the ground truth 3D model.}
	\label{figure:traj}
\end{figure*}

\textbf{Qualitative results.} In Figure~\ref{figure:traj}, we visualize the camera poses on the 7-Scenes dataset. We are glad to see that the estimated poses overall overlap their ground truth in the challenging few-shot setting. However, for the regions which are far from training viewpoints, our method shows large jitters in pose estimation. Following recent works~\cite{li2020hierarchical, brachmann2021dsacstar}, we also render the color images with the estimated poses for intuitive comparisons. The rendered and query images are well aligned, which validates the localization accuracy of our method. Please refer to the supplementary for the visual results on Cambridge landmarks. 

\subsection{Analysis} \label{subsec:analysis}
\begin{figure}[tbh]
\centering
	\includegraphics[width=0.9\linewidth]{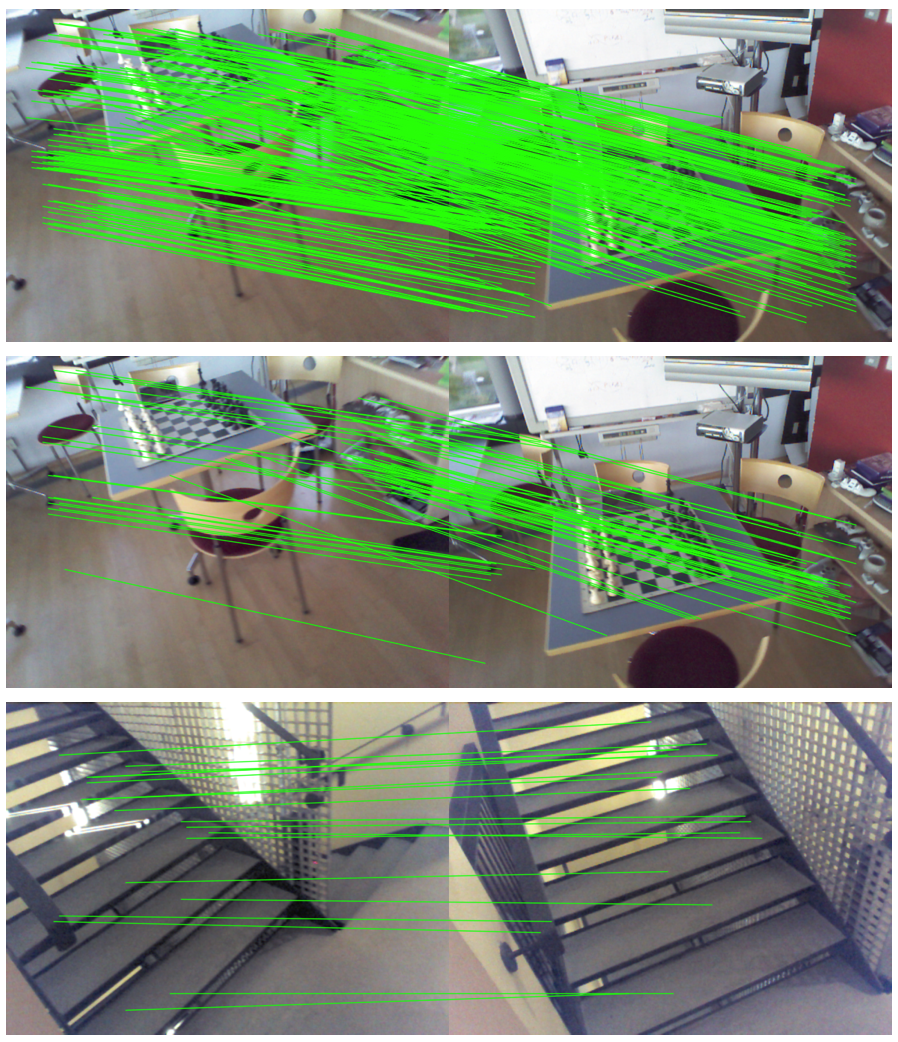}
	\caption{We visualize the correct matches. The correctness is determined by their ground truth coordinates with $5$cm tolerance. Top: the model is trained with original training images and tested in the same scene. Middle: the model is trained with few-shot images and tested in the same scene. Bottom: the model is trained with original training images and tested in a novel scene.  }
	\label{figure:matching}
\end{figure}

\textbf{Scene coordinate regression.} 
This paragraph provides an analysis of a typical scene coordinate regression network, SCRNet~\cite{li2020hierarchical}, to further explain our inspiration. 
Our hypothesis is that the first-to-middle layers of SCRNet serve as a feature extractor for robust visual descriptors under different viewpoints while the rest layers memorize the mapping from descriptors to scene-specific coordinates. 
The features from the former become less effective when the number of training data decreases. 
To validate this hypothesis, we conduct feature matching experiments using the intermediate feature map of SCRNet.
Three examples are shown in Figure~\ref{figure:matching},  where we train the model on \textsc{Chess} and employ the middle layer's output feature map as visual descriptors.
The top row shows that when trained with full training images it obtains hundreds of correct matches. 
When the model is trained with only few-shot images, we observe a significant decrease in the correct matches, as shown in the middle row, which indicates that the visual descriptors require amounts of training data. 
On the bottom row, if the model is applied to extract features in a novel scene \textsc{Stairs}, the number of correct matches is further decreased. 
These experiments demonstrate that although the first-to-middle layers of SCRNet have the ability to extract visual descriptors, they require huge training data and hardly be generalized to novel scenes. 
Consequently, we adopt the decoupled hypothesis of scene coordinate regression and employ an off-the-shelf feature extractor SuperPoint to handle the few-shot setting. 

\textbf{Off-the-shelf feature extractors.} We apply SuperPoint~\cite{detone2018superpoint} as our default feature extractor in Table~\ref{tab:main1} and Table~\ref{tab:main2}. The results in Table~\ref{tab:features} demonstrate that the extractor can also be replaced by other visual descriptors. We apply two state-of-the-art learned features D2-Net~\cite{dusmanu2019d2} and R2D2~\cite{revaud2019r2d2} as alternatives. Similar to SuperPoint, for the two methods, we take only their semi-dense visual descriptors without detectors. Since these learned features have different dimensions, we slightly change the input channels of the followed classification network to align with different visual descriptors. The results show that the different feature extractors have similar performance in pose estimation.
\begin{table}[t!]
\centering
\resizebox{.38\textwidth}{!}{
	\begin{tabular}{l | c c | c c  }
	\toprule
		 \multirow{2}{*}{Scene} & \multicolumn{2}{c}{D2-Net \cite{dusmanu2019d2}} & \multicolumn{2}{c}{R2D2 \cite{revaud2019r2d2}} \\
 		 & $t$ (cm) & $r$ ($^\circ$) & $t$ (cm) & $r$ ($^\circ$) \\
		\midrule\midrule
 		\textsc{Chess}      & 4 & 1.20 & 5 & 1.57 \\
 		\textsc{Fire}       & 4 & 1.46 & 4 & 1.55 \\
 		\textsc{Heads}      & 2 & 1.44 & 2 & 1.50 \\
 	    \textsc{Office}     & 5 & 1.52 & 6 & 1.72 \\
 		\textsc{Pumpkin}    & 7 & 1.69 & 6 & 1.48 \\
 		\textsc{RedKitchen} & 6 & 1.93 & 7 & 1.89 \\
 		\textsc{Stairs}     & 7 & 1.96 & 7 & 1.95 \\
    \bottomrule
	\end{tabular}
}
\vspace{10pt}
\caption{Impact of different feature extractors. We replace our feature extractor with state-of-the-art alternatives and evaluate these variants on the 7-Scenes dataset. The median errors of estimated camera poses are comparable with the results in Table~\ref{tab:main1}. }
\label{tab:features}
\end{table}

\textbf{Scene region partitions.} 
To explore the impact of different scene region partitions, we train and evaluate our model with a different number of region classes. We perform experiments on two selected scenes, \textsc{RedKitchen} for indoor and \textsc{GreatCourt} for outdoor. 
As shown in Table~\ref{tab:classes}, we observe that more classes result in more accurate camera pose estimation. 
The results from \textsc{RedKitchen} indicate that the effectiveness of the finer region partition is limited since the current regions are sufficient for the pose estimation. 
The improvements are significant in \textsc{GreatCourt}. 
Especially, when the regions increase to $120 \times 120$, the translation error decreases to $72$cm, which is more than $40$\% lower than $124$cm. However, the fine-grained region partition suffers from heavy computation costs, and the processing time significantly exceeds our training time with 14400 regions.

\begin{table}[t!]
\centering
\resizebox{.45\textwidth}{!}{
	\begin{tabular}{l | c c | c c  }
	\toprule
		 \multirow{2}{*}{\# Classes} & \multicolumn{2}{c}{\textsc{RedKitchen}} & \multicolumn{2}{c}{\textsc{GreatCourt}} \\
 		 & $t$ (cm) & $r$ ($^\circ$) & $t$ (cm) & $r$ ($^\circ$) \\
		\midrule\midrule
		$30 \times 30 = 900$ & 7 & 2.09  & - & - \\
		$50 \times 50 = 2500$ & 6 & 1.92  & 124 & 0.68 \\
		$80 \times 80 = 6400$ & 6 & 1.90  & 101 & 0.58 \\
		$90 \times 90 = 8100$ & 6 & 1.85  & 99 & 0.56 \\
		$120 \times 120 = 14400$ & - & -  & 72 & 0.40 \\
    \bottomrule
	\end{tabular}
}
\vspace{10pt}
\caption{Impact of different scene region partitions. We report the median errors of estimated camera poses in two scenes from the 7-Scenes dataset and Cambridge landmarks. }
\label{tab:classes}
\end{table}

\begin{figure}[tbh]
\centering
	\includegraphics[width=1\linewidth]{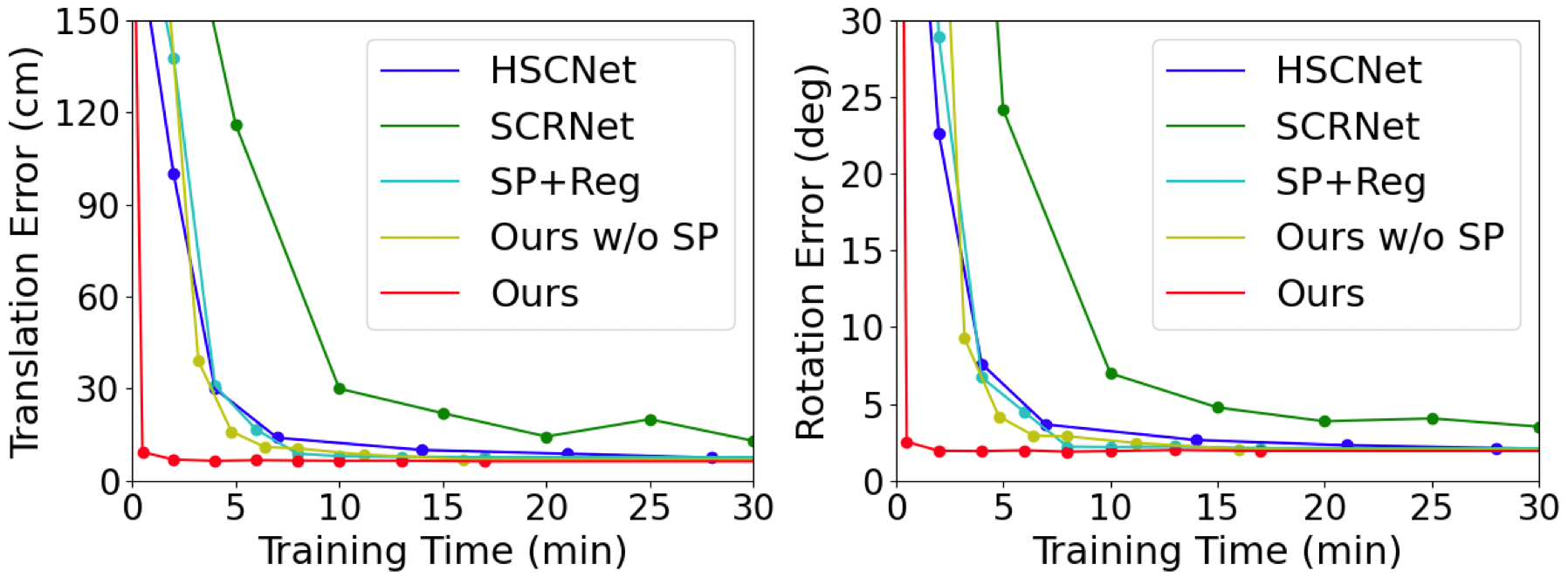}
	\vspace{-2mm}
	\caption{
	The camera pose median errors vs. training time.
	}
	\label{figure:time-pose}
\end{figure}

\textbf{Training times.} The key feature of our method is the fast memorization ability with few-shot images. As shown in Figure~\ref{figure:time-pose}, we compare our training time with SCRNet, HSCNet, and several baselines in \textsc{RedKitchen}. For fair comparisons, we train our classifier from random initialization without the meta-learning based pre-training. The results show that our method achieves the fastest convergence with only a few minutes of training time. The baseline named Ours w/o SP denotes a randomly initialized feature extractor instead of SuperPoint, and the comparison with it demonstrates the effectiveness of pre-learned visual descriptors. As we expect, SCRNet is the slowest to converge. Applying SuperPoint to SCR (i.e. SP+Reg) significantly accelerates the convergence and makes it approach HSCNet. HSCNet achieves a not-bad convergence speed as it also applies hierarchical classifications besides coordinate regression. 
To conclude, these experiments validate the effectiveness of our decoupled design with scene region classification. 

\begin{figure}[tbh]
\centering
	\includegraphics[width=1\linewidth]{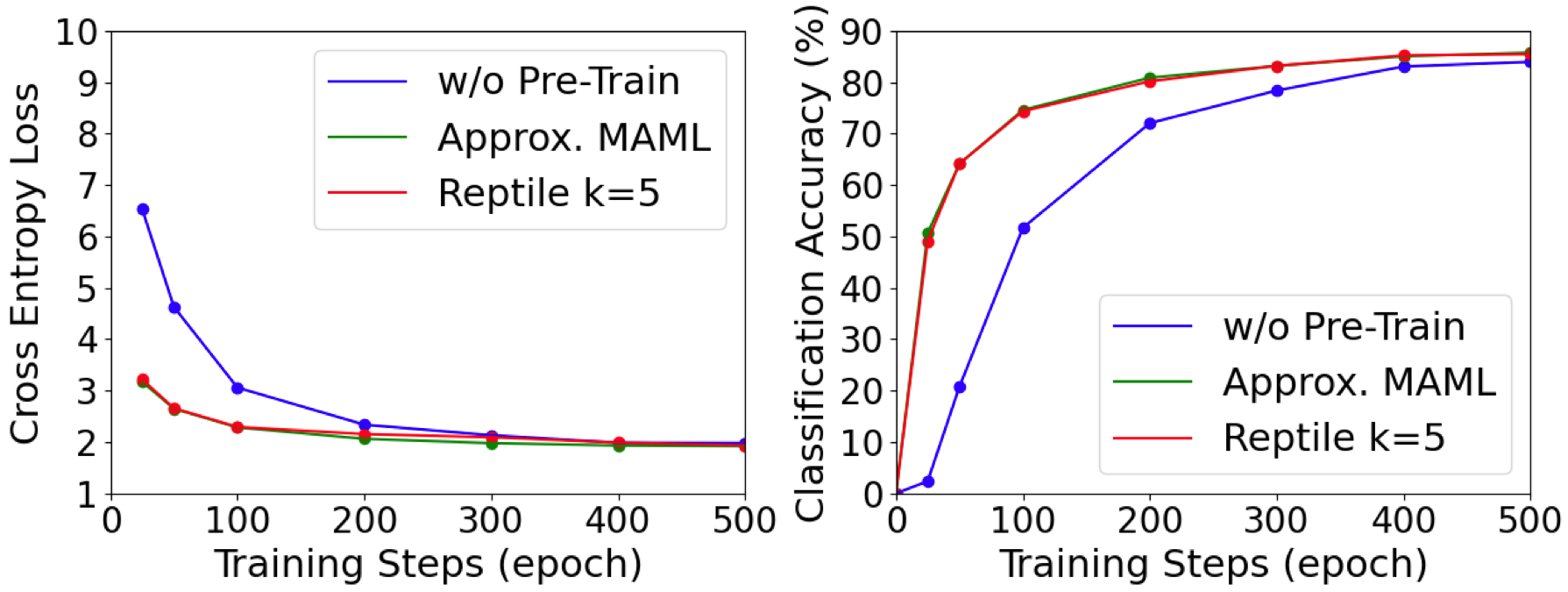}
	\vspace{-2mm}
	\caption{
	Our training curves with and without meta-learning. 
	}
	\label{figure:meta}
\end{figure}

\textbf{Effectiveness of meta-learning.} 
In Figure~\ref{figure:meta}, we report the training curves of our method in \textsc{RedKitchen}. 
We compare three setups: training from random initialization, from pre-trained models by Reptile with $k=2$ (approximation of MAML\cite{finn2017model}) and $k=5$. We observe that the pre-training accelerates the training convergence, while different types of Reptile gradients do not make much difference.

\section{Conclusion}
In this paper, we propose a novel problem setting that performs visual localization with only few-shot posed images as the database, along with a simple and effective method based on hierarchical scene region classification. Experiments and analysis validate the design of our method, which achieves fast scene memorization with low localization error. Despite the high efficiency, our method suffers from a performance drop due to the few-shot setting. Improving the camera pose accuracy worth more explorations in future works.

\textbf{Acknowledgements.} 
We thank Songyin Wu, Shanghang Zhang, and Kai Xu for their early discussions and Shaohui Liu for data processing. 
This work was supported in part by NSFC Projects of International Cooperation and Exchanges (62161146002), NSFC (61902007), Academic of Finland (grant 327911) and Aalto Science-IT project. 
Siyan Dong was supported by the China Scholarship Council.

\section*{Appendix}
\appendix

In this appendix, we report the details of our few-shot version of datsets 7-Scenes~\cite{shotton2013scene} and Cambridge~\cite{kendall2015posenet} in section~\ref{sec:data}, the additional results of SCRNet~\cite{li2020hierarchical} based feature matching in section~\ref{sec:match}, the qualitative results on Cambridge in section~\ref{sec:vis}, the efficiency of our method in section~\ref{sec:time}, and the impact of leaf granularity in section~\ref{sec:granularity}.

\begin{figure*}[hbt]
\centering
	\includegraphics[width=.98\linewidth]{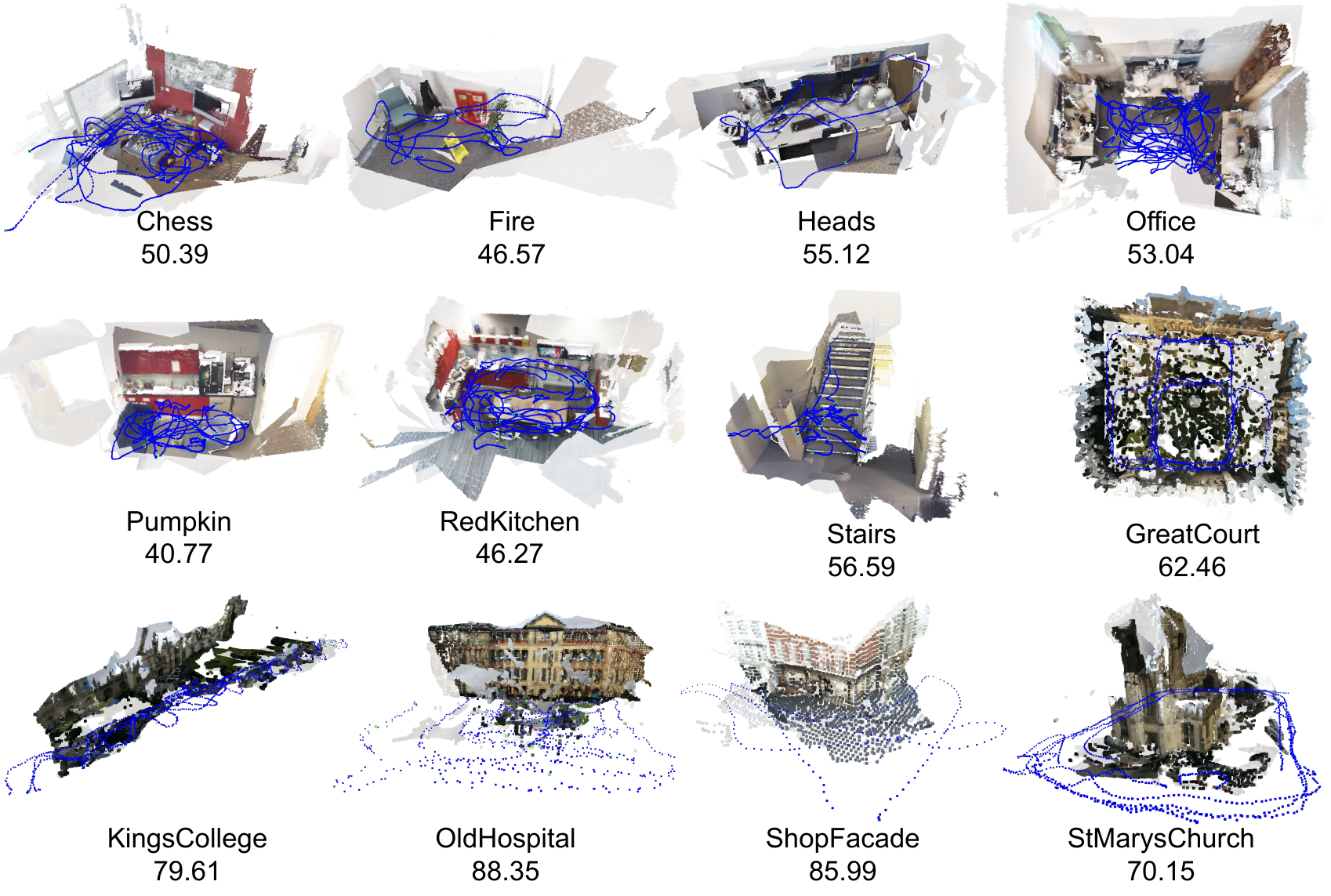}
	\caption{Visualization of the original (blue dots) and the few-shot (green dots) training sets. The reconstruction from the few-shot set is shown in colors while the full scene model is shown as the background. 
	We report the area coverage (\%) of the few-shot reconstruction for each scene.
	The coverage is computed with 10cm and 50cm tolerances for 7-Scenes and Cambridge, respectively. }
	\label{figure:stat}
\end{figure*}

\begin{figure*}[htb!]
\centering
	\includegraphics[width=.98\linewidth]{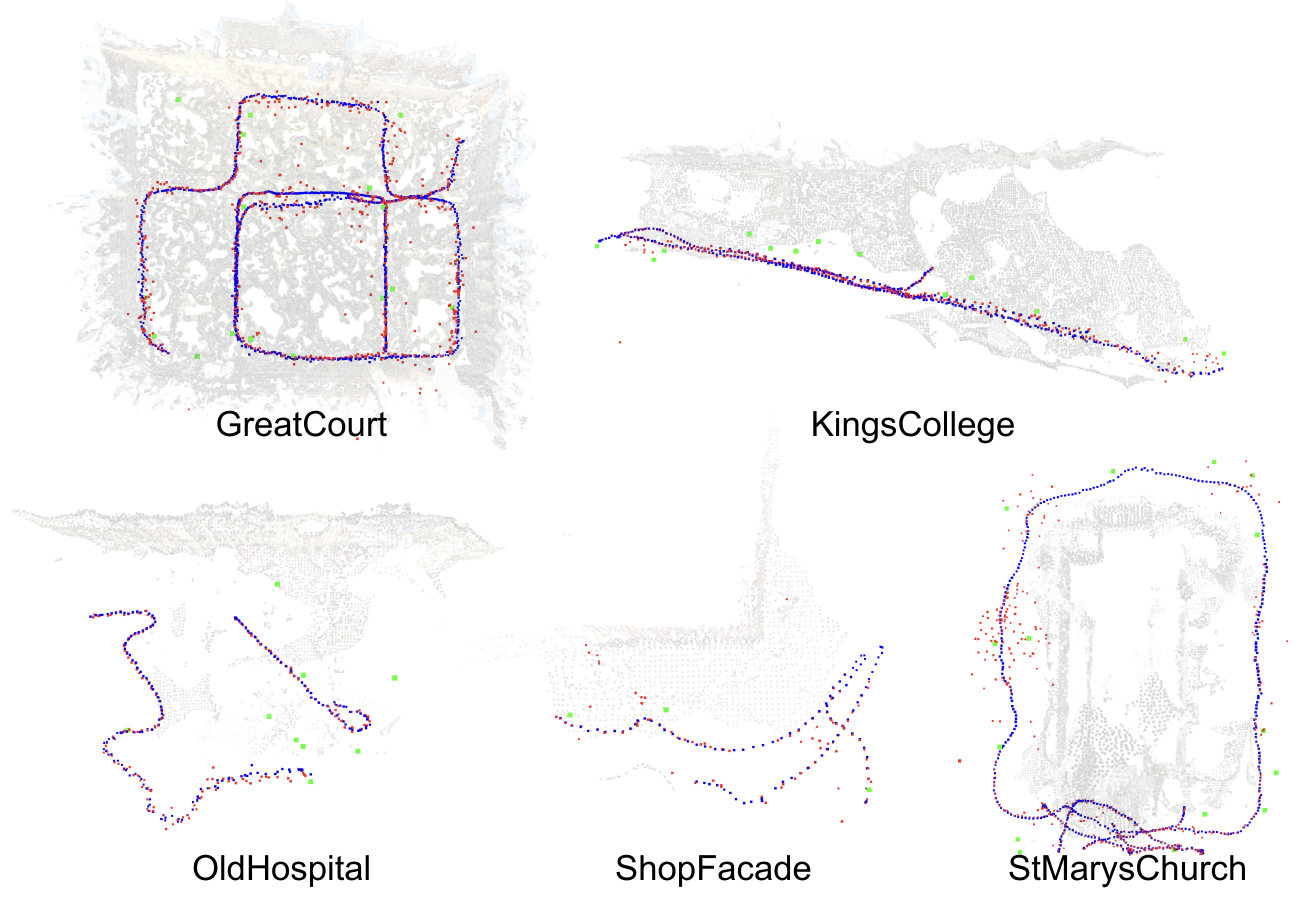}
	\caption{Visual results on the Cambridge landmarks. 
	Training images (green dots), test images (blue dots) and our estimates (red dots).
	}
	\label{figure:vis_cambridge}
\end{figure*}

\def\thesection{\Alph{section}}

\section{Few-Shot Datasets} \label{sec:data}
We perform few-shot experiments on the 7-Scenes dataset and the Cambridge landmarks in the main paper. To avoid bias, the few-shot training set is uniformly sampled  from the original training set, as shown in Figure~\ref{figure:stat}.

\section{SCRNet Based Feature Matching} \label{sec:match}
The main paper showcases examples that feature maps from SCRNet can match amounts of correct points in the same scene, and match fewer points in another scene. In this section, we perform a cross validation and report quantitative results. As shown in Table~\ref{tab:cross}, we build image pairs by skipping every 30 images, and we count the average number of correct matches. We observe that the test results in another scene achieve consistently less correct matches than those in the same scene. Note that \textsc{Stairs} is a very challenging scene with repetitive patterns and textureless regions, therefore the numbers of correct matches are overall less than those in \textsc{Chess}.

\begin{table}[tbh]
\centering
\resizebox{.25\textwidth}{!}{
	\begin{tabular}{l | c c }
	\toprule
		 \multirow{2}{*}{Test scene} & \multicolumn{2}{c}{Training scene} \\
		 & \textsc{Chess} & \textsc{Stairs} \\
		 \midrule
		 \textsc{Chess} & 746 & 564 \\
		 \textsc{Stairs} & 336 & 486 \\
    \bottomrule
	\end{tabular}
}
\vspace{5pt}
\caption{The number of correct matches. The model is trained separately in \textsc{Chess} and \textsc{Stairs} and tested in the same scene and cross scenes. }
\label{tab:cross}
\end{table}

\section{Qualitative Results on Cambridge} \label{sec:vis}
In the main paper, we show the qualitative results on the 7-Scenes dataset. The qualitative reuslts on the Cambridge landmarks are shown in Figure~\ref{figure:vis_cambridge}. We observe that in overall cases, the estimated poses overlap their ground truth.

\section{Time Consumption} \label{sec:time}
In this section, we provide detailed time consumption. 
Our method takes $\sim$63 ms for each training iteration (SCRNet $\sim$100 ms, and HSCNet~\cite{li2020hierarchical} $\sim$125 ms). The meta-learning based pre-training takes about 5 hours and is once for all. The few-shot memorization converges with about 2 minutes. For inference, our method runs ($\sim$~200 ms) slower than scene coordinate regression based methods (SCRNet $\sim$130 ms) and faster than feature matching based methods (HLoc~\cite{sarlin2019coarse, sarlin2020superglue} $\sim$~500 ms). This is because of the more complexity of the one-to-many PnP-RANSAC mechanism.

\section{Leaf Granularity} \label{sec:granularity}
We report the running time and pose accuracy using different granularities of clustering in leaf nodes shown in Table~\ref{tab:cluster}. The choice of granularity is a trade-off between accuracy and efficiency. In general, the accuracy improves when using finer-grained granularities. However, an overly high cluster number, e.g. $q=50$, may causes more randomness for RANSAC, leading to performance reduction.

\begin{table}[htb]
\centering
\resizebox{.45\textwidth}{!}{
\begin{tabular}{l|c|c|c|c}
\hline
\# Clusters & $q=1$ & $q=10$ & $q=30$ & $q=50$ \\ 
\hline

RANSAC time (ms) & 20-50 & 80-230 & 180-500 & 400-700 \\ 

Median error (cm / $^\circ$) & 8/2.00 & 6/1.93 & 6/1.89 & 6/1.95 \\ 
\hline
\end{tabular}
}
\vspace{5pt}
\caption{Impact of the number of leaf node clusters. Experiments are conducted in the scene \textsc{RedKitchen}. }
\label{tab:cluster}
\end{table}

{\small
\bibliographystyle{ieee_fullname}
\bibliography{egbib}

\begin{thebibliography}{10}\itemsep=-1pt

\bibitem{agarap2018deep}
Abien~Fred Agarap.
\newblock Deep learning using rectified linear units (relu).
\newblock {\em arXiv preprint arXiv:1803.08375}, 2018.

\bibitem{andrychowicz2016learning}
Marcin Andrychowicz, Misha Denil, Sergio Gomez, Matthew~W Hoffman, David Pfau,
  Tom Schaul, Brendan Shillingford, and Nando De~Freitas.
\newblock Learning to learn by gradient descent by gradient descent.
\newblock {\em Advances in neural information processing systems}, 29, 2016.

\bibitem{antoniou2017data}
Antreas Antoniou, Amos Storkey, and Harrison Edwards.
\newblock Data augmentation generative adversarial networks.
\newblock {\em arXiv preprint arXiv:1711.04340}, 2017.

\bibitem{arandjelovic2016netvlad}
Relja Arandjelovic, Petr Gronat, Akihiko Torii, Tomas Pajdla, and Josef Sivic.
\newblock Netvlad: Cnn architecture for weakly supervised place recognition.
\newblock In {\em Proceedings of the IEEE conference on computer vision and
  pattern recognition}, pages 5297--5307, 2016.

\bibitem{arandjelovic2013all}
Relja Arandjelovic and Andrew Zisserman.
\newblock All about vlad.
\newblock In {\em Proceedings of the IEEE conference on Computer Vision and
  Pattern Recognition}, pages 1578--1585, 2013.

\bibitem{ba2016layer}
Jimmy~Lei Ba, Jamie~Ryan Kiros, and Geoffrey~E Hinton.
\newblock Layer normalization.
\newblock {\em arXiv preprint arXiv:1607.06450}, 2016.

\bibitem{brachmann2017dsac}
Eric Brachmann, Alexander Krull, Sebastian Nowozin, Jamie Shotton, Frank
  Michel, Stefan Gumhold, and Carsten Rother.
\newblock {DSAC}-{Differentiable RANSAC} for camera localization.
\newblock In {\em CVPR}, 2017.

\bibitem{brachmann2018lessmore}
Eric Brachmann and Carsten Rother.
\newblock Learning less is more - {6D} camera localization via {3D} surface
  regression.
\newblock In {\em CVPR}, 2018.

\bibitem{brachmann2021dsacstar}
Eric Brachmann and Carsten Rother.
\newblock Visual camera re-localization from {RGB} and {RGB-D} images using
  {DSAC}.
\newblock {\em TPAMI}, 2021.

\bibitem{cavallari2017fly}
Tommaso Cavallari, Stuart Golodetz, Nicholas~A Lord, Julien Valentin, Luigi
  Di~Stefano, and Philip~HS Torr.
\newblock On-the-fly adaptation of regression forests for online camera
  relocalisation.
\newblock In {\em Proceedings of the IEEE conference on computer vision and
  pattern recognition}, pages 4457--4466, 2017.

\bibitem{cavallari2019real}
Tommaso Cavallari, Stuart Golodetz, Nicholas~A Lord, Julien Valentin, Victor~A
  Prisacariu, Luigi Di~Stefano, and Philip~HS Torr.
\newblock Real-time rgb-d camera pose estimation in novel scenes using a
  relocalisation cascade.
\newblock {\em IEEE transactions on pattern analysis and machine intelligence},
  42(10):2465--2477, 2019.

\bibitem{dai2017bundlefusion}
Angela Dai, Matthias Nie{\ss}ner, Michael Zollh{\"o}fer, Shahram Izadi, and
  Christian Theobalt.
\newblock Bundlefusion: Real-time globally consistent 3d reconstruction using
  on-the-fly surface reintegration.
\newblock {\em ACM Transactions on Graphics (ToG)}, 36(4):1, 2017.

\bibitem{detone2018superpoint}
Daniel DeTone, Tomasz Malisiewicz, and Andrew Rabinovich.
\newblock Superpoint: Self-supervised interest point detection and description.
\newblock In {\em Proceedings of the IEEE conference on computer vision and
  pattern recognition workshops}, pages 224--236, 2018.

\bibitem{dong2021robust}
Siyan Dong, Qingnan Fan, He Wang, Ji Shi, Li Yi, Thomas Funkhouser, Baoquan
  Chen, and Leonidas~J Guibas.
\newblock Robust neural routing through space partitions for camera
  relocalization in dynamic indoor environments.
\newblock In {\em Proceedings of the IEEE/CVF Conference on Computer Vision and
  Pattern Recognition}, pages 8544--8554, 2021.

\bibitem{dusmanu2019d2}
Mihai Dusmanu, Ignacio Rocco, Tomas Pajdla, Marc Pollefeys, Josef Sivic,
  Akihiko Torii, and Torsten Sattler.
\newblock D2-net: A trainable cnn for joint description and detection of local
  features.
\newblock In {\em Proceedings of the IEEE/cvf conference on computer vision and
  pattern recognition}, pages 8092--8101, 2019.

\bibitem{eade2013gauss}
Ethan Eade.
\newblock Gauss-newton/levenberg-marquardt optimization.
\newblock {\em Tech. Rep.}, 2013.

\bibitem{finn2017model}
Chelsea Finn, Pieter Abbeel, and Sergey Levine.
\newblock Model-agnostic meta-learning for fast adaptation of deep networks.
\newblock In {\em International conference on machine learning}, pages
  1126--1135. PMLR, 2017.

\bibitem{fischler1981random}
Martin~A Fischler and Robert~C Bolles.
\newblock Random sample consensus: a paradigm for model fitting with
  applications to image analysis and automated cartography.
\newblock {\em Communications of the ACM}, 24(6):381--395, 1981.

\bibitem{gao2003complete}
Xiao-Shan Gao, Xiao-Rong Hou, Jianliang Tang, and Hang-Fei Cheng.
\newblock Complete solution classification for the perspective-three-point
  problem.
\newblock {\em IEEE transactions on pattern analysis and machine intelligence},
  25(8):930--943, 2003.

\bibitem{glocker2014real}
Ben Glocker, Jamie Shotton, Antonio Criminisi, and Shahram Izadi.
\newblock Real-time rgb-d camera relocalization via randomized ferns for
  keyframe encoding.
\newblock {\em IEEE transactions on visualization and computer graphics},
  21(5):571--583, 2014.

\bibitem{hariharan2017low}
Bharath Hariharan and Ross Girshick.
\newblock Low-shot visual recognition by shrinking and hallucinating features.
\newblock In {\em Proceedings of the IEEE International Conference on Computer
  Vision}, pages 3018--3027, 2017.

\bibitem{huang2021vs}
Zhaoyang Huang, Han Zhou, Yijin Li, Bangbang Yang, Yan Xu, Xiaowei Zhou, Hujun
  Bao, Guofeng Zhang, and Hongsheng Li.
\newblock Vs-net: Voting with segmentation for visual localization.
\newblock In {\em Proceedings of the IEEE/CVF Conference on Computer Vision and
  Pattern Recognition}, pages 6101--6111, 2021.

\bibitem{ke2017efficient}
Tong Ke and Stergios~I Roumeliotis.
\newblock An efficient algebraic solution to the perspective-three-point
  problem.
\newblock In {\em Proceedings of the IEEE Conference on Computer Vision and
  Pattern Recognition}, pages 7225--7233, 2017.

\bibitem{Kendall_2017_CVPR}
Alex Kendall and Roberto Cipolla.
\newblock Geometric loss functions for camera pose regression with deep
  learning.
\newblock In {\em Proceedings of the IEEE Conference on Computer Vision and
  Pattern Recognition (CVPR)}, July 2017.

\bibitem{kendall2015posenet}
Alex Kendall, Matthew Grimes, and Roberto Cipolla.
\newblock Posenet: A convolutional network for real-time 6-dof camera
  relocalization.
\newblock In {\em Proceedings of the IEEE international conference on computer
  vision}, pages 2938--2946, 2015.

\bibitem{kingma2014adam}
Diederik~P Kingma and Jimmy Ba.
\newblock Adam: A method for stochastic optimization.
\newblock {\em arXiv preprint arXiv:1412.6980}, 2014.

\bibitem{koch2015siamese}
Gregory Koch, Richard Zemel, Ruslan Salakhutdinov, et~al.
\newblock Siamese neural networks for one-shot image recognition.
\newblock In {\em ICML deep learning workshop}, volume~2, page~0. Lille, 2015.

\bibitem{li2020hierarchical}
Xiaotian Li, Shuzhe Wang, Yi Zhao, Jakob Verbeek, and Juho Kannala.
\newblock Hierarchical scene coordinate classification and regression for
  visual localization.
\newblock In {\em Proceedings of the IEEE/CVF Conference on Computer Vision and
  Pattern Recognition}, pages 11983--11992, 2020.

\bibitem{li2018full}
Xiaotian Li, Juha Ylioinas, and Juho Kannala.
\newblock Full-frame scene coordinate regression for image-based localization.
\newblock {\em arXiv preprint arXiv:1802.03237}, 2018.

\bibitem{nichol2018reptile}
Alex Nichol and John Schulman.
\newblock Reptile: a scalable metalearning algorithm.
\newblock {\em arXiv preprint arXiv:1803.02999}, 2(3):4, 2018.

\bibitem{scikit-learn}
F. Pedregosa, G. Varoquaux, A. Gramfort, V. Michel, B. Thirion, O. Grisel, M.
  Blondel, P. Prettenhofer, R. Weiss, V. Dubourg, J. Vanderplas, A. Passos, D.
  Cournapeau, M. Brucher, M. Perrot, and E. Duchesnay.
\newblock Scikit-learn: Machine learning in {P}ython.
\newblock {\em Journal of Machine Learning Research}, 12:2825--2830, 2011.

\bibitem{revaud2019r2d2}
Jerome Revaud, Cesar De~Souza, Martin Humenberger, and Philippe Weinzaepfel.
\newblock R2d2: Reliable and repeatable detector and descriptor.
\newblock {\em Advances in neural information processing systems}, 32, 2019.

\bibitem{sarlin2019coarse}
Paul-Edouard Sarlin, Cesar Cadena, Roland Siegwart, and Marcin Dymczyk.
\newblock From coarse to fine: Robust hierarchical localization at large scale.
\newblock In {\em CVPR}, 2019.

\bibitem{sarlin2020superglue}
Paul-Edouard Sarlin, Daniel DeTone, Tomasz Malisiewicz, and Andrew Rabinovich.
\newblock {SuperGlue}: Learning feature matching with graph neural networks.
\newblock In {\em CVPR}, 2020.

\bibitem{sarlin2021back}
Paul-Edouard Sarlin, Ajaykumar Unagar, Mans Larsson, Hugo Germain, Carl Toft,
  Viktor Larsson, Marc Pollefeys, Vincent Lepetit, Lars Hammarstrand, Fredrik
  Kahl, et~al.
\newblock Back to the feature: Learning robust camera localization from pixels
  to pose.
\newblock In {\em Proceedings of the IEEE/CVF Conference on Computer Vision and
  Pattern Recognition}, pages 3247--3257, 2021.

\bibitem{sattler2016efficient}
Torsten Sattler, Bastian Leibe, and Leif Kobbelt.
\newblock Efficient \& effective prioritized matching for large-scale
  image-based localization.
\newblock {\em IEEE transactions on pattern analysis and machine intelligence},
  39(9):1744--1756, 2016.

\bibitem{sattler2012image}
Torsten Sattler, Tobias Weyand, Bastian Leibe, and Leif Kobbelt.
\newblock Image retrieval for image-based localization revisited.
\newblock In {\em BMVC}, volume~1, page~4, 2012.

\bibitem{sattler2019understanding}
Torsten Sattler, Qunjie Zhou, Marc Pollefeys, and Laura Leal-Taixe.
\newblock Understanding the limitations of cnn-based absolute camera pose
  regression.
\newblock In {\em Proceedings of the IEEE/CVF conference on computer vision and
  pattern recognition}, pages 3302--3312, 2019.

\bibitem{shotton2013scene}
Jamie Shotton, Ben Glocker, Christopher Zach, Shahram Izadi, Antonio Criminisi,
  and Andrew Fitzgibbon.
\newblock Scene coordinate regression forests for camera relocalization in
  rgb-d images.
\newblock In {\em Proceedings of the IEEE Conference on Computer Vision and
  Pattern Recognition}, pages 2930--2937, 2013.

\bibitem{simonyan2014very}
Karen Simonyan and Andrew Zisserman.
\newblock Very deep convolutional networks for large-scale image recognition.
\newblock {\em arXiv preprint arXiv:1409.1556}, 2014.

\bibitem{snell2017prototypical}
Jake Snell, Kevin Swersky, and Richard Zemel.
\newblock Prototypical networks for few-shot learning.
\newblock {\em Advances in neural information processing systems}, 30, 2017.

\bibitem{sun2021loftr}
Jiaming Sun, Zehong Shen, Yuang Wang, Hujun Bao, and Xiaowei Zhou.
\newblock Loftr: Detector-free local feature matching with transformers.
\newblock In {\em Proceedings of the IEEE/CVF conference on computer vision and
  pattern recognition}, pages 8922--8931, 2021.

\bibitem{tang2021learning}
Shitao Tang, Chengzhou Tang, Rui Huang, Siyu Zhu, and Ping Tan.
\newblock Learning camera localization via dense scene matching.
\newblock In {\em Proceedings of the IEEE/CVF Conference on Computer Vision and
  Pattern Recognition}, pages 1831--1841, 2021.

\bibitem{torii201524}
Akihiko Torii, Relja Arandjelovic, Josef Sivic, Masatoshi Okutomi, and Tomas
  Pajdla.
\newblock 24/7 place recognition by view synthesis.
\newblock In {\em Proceedings of the IEEE conference on computer vision and
  pattern recognition}, pages 1808--1817, 2015.

\bibitem{uy2018pointnetvlad}
Mikaela~Angelina Uy and Gim~Hee Lee.
\newblock Pointnetvlad: Deep point cloud based retrieval for large-scale place
  recognition.
\newblock In {\em Proceedings of the IEEE conference on computer vision and
  pattern recognition}, pages 4470--4479, 2018.

\bibitem{valentin2016learning}
Julien Valentin, Angela Dai, Matthias Nie{\ss}ner, Pushmeet Kohli, Philip Torr,
  Shahram Izadi, and Cem Keskin.
\newblock Learning to navigate the energy landscape.
\newblock In {\em 2016 Fourth International Conference on 3D Vision (3DV)},
  pages 323--332. IEEE, 2016.

\bibitem{valentin2015exploiting}
Julien Valentin, Matthias Nie{\ss}ner, Jamie Shotton, Andrew Fitzgibbon,
  Shahram Izadi, and Philip~HS Torr.
\newblock Exploiting uncertainty in regression forests for accurate camera
  relocalization.
\newblock In {\em Proceedings of the IEEE conference on computer vision and
  pattern recognition}, pages 4400--4408, 2015.

\bibitem{vinyals2016matching}
Oriol Vinyals, Charles Blundell, Timothy Lillicrap, Daan Wierstra, et~al.
\newblock Matching networks for one shot learning.
\newblock {\em Advances in neural information processing systems}, 29, 2016.

\bibitem{walch2017image}
Florian Walch, Caner Hazirbas, Laura Leal-Taixe, Torsten Sattler, Sebastian
  Hilsenbeck, and Daniel Cremers.
\newblock Image-based localization using lstms for structured feature
  correlation.
\newblock In {\em Proceedings of the IEEE International Conference on Computer
  Vision}, pages 627--637, 2017.

\bibitem{wang2020atloc}
Bing Wang, Changhao Chen, Chris~Xiaoxuan Lu, Peijun Zhao, Niki Trigoni, and
  Andrew Markham.
\newblock Atloc: Attention guided camera localization.
\newblock In {\em Proceedings of the AAAI Conference on Artificial
  Intelligence}, volume~34, pages 10393--10401, 2020.

\bibitem{wang2021continual}
Shuzhe Wang, Zakaria Laskar, Iaroslav Melekhov, Xiaotian Li, and Juho Kannala.
\newblock Continual learning for image-based camera localization.
\newblock In {\em Proceedings of the IEEE/CVF International Conference on
  Computer Vision}, pages 3252--3262, 2021.

\bibitem{wang2018low}
Yu-Xiong Wang, Ross Girshick, Martial Hebert, and Bharath Hariharan.
\newblock Low-shot learning from imaginary data.
\newblock In {\em Proceedings of the IEEE conference on computer vision and
  pattern recognition}, pages 7278--7286, 2018.

\bibitem{wu2018group}
Yuxin Wu and Kaiming He.
\newblock Group normalization.
\newblock In {\em Proceedings of the European conference on computer vision
  (ECCV)}, pages 3--19, 2018.

\bibitem{yang2019sanet}
Luwei Yang, Ziqian Bai, Chengzhou Tang, Honghua Li, Yasutaka Furukawa, and Ping
  Tan.
\newblock Sanet: Scene agnostic network for camera localization.
\newblock In {\em Proceedings of the IEEE/CVF International Conference on
  Computer Vision}, pages 42--51, 2019.

\end{thebibliography}
}

\end{document}